\pdfoutput=1

\documentclass[11pt]{article}

\usepackage[]{EMNLP2023}

\usepackage{times}
\usepackage{latexsym}
\usepackage{amsmath}
\usepackage{tabularx}

\usepackage{booktabs}
\usepackage{rotating}
\usepackage{xcolor,colortbl}
\usepackage{CJKutf8}

\usepackage[T1]{fontenc}

\usepackage[utf8]{inputenc}

\usepackage{microtype}

\usepackage{inconsolata}

\usepackage{pifont}
\definecolor{darkpastelgreen}{rgb}{0.01, 0.75, 0.24}
\definecolor{darkpastelred}{rgb}{0.76, 0.23, 0.13}
\newcommand{\cmark}{\textcolor{darkpastelgreen}{\ding{51}}}%
\newcommand{\xmark}{\textcolor{darkpastelred}{\ding{55}}}%

\title{ACES: Translation Accuracy Challenge Sets at WMT 2023}

\author{Chantal Amrhein$^{1,2}$\thanks{{ } Equal contribution by all authors.} \and Nikita Moghe$^{3}$\footnotemark[1]  \and Liane Guillou$^{4}$\footnotemark[1] \\
  $^1$Textshuttle, Zurich \\
  $^2$Department of Computational Linguistics, University of Zurich\\
  $^3$School of Informatics, University of Edinburgh \\
  $^4$Department of Computer Science, RISE Research Institutes of Sweden \\\medskip
  \texttt{amrhein@textshuttle.com, nikita.moghe@ed.ac.uk, liane.guillou@ri.se}}

\begin{document}
\maketitle
\begin{abstract}
We benchmark the performance of segment-level metrics submitted to WMT 2023 using the \textsc{ACES} Challenge Set \citep{amrhein-etal-2022-aces}. The challenge set consists of 36K examples representing challenges from 68 phenomena and covering 146 language pairs. The phenomena range from simple perturbations at the word/character level to more complex errors based on discourse and real-world knowledge. For each metric, we provide a detailed profile of performance over a range of error categories as well as an overall \textit{ACES}-Score for quick comparison. We also measure the incremental performance of the metrics submitted to both WMT 2023 and 2022. 
We find that 1) there is no clear \textit{winner} among the metrics submitted to WMT 2023, and 2) performance change between the 2023 and 2022 versions of the metrics is highly variable. Our recommendations are similar to those from WMT 2022. Metric developers should focus on: building ensembles of metrics from different design families, developing metrics that pay more attention to the source and rely less on surface-level overlap, and carefully determining the influence of multilingual embeddings on MT evaluation. 
\end{abstract}

\section{Introduction}
\label{sec:introduction}
Challenge sets are a useful tool in measuring the performance of systems or metrics on one or more specific phenomena of interest. They may be used to compare the performance of a range of \textit{different} systems or to identify performance improvement/degradation between successive iterations of the \textit{same} system. 

Challenge sets exist for a range of natural language processing (NLP) tasks including Sentiment Analysis \citep{li-etal-2017-bibi,mahler-etal-2017-breaking,staliunaite-bonfil-2017-breaking}, Natural Language Inference \citep{mccoy2019non,Rocchietti2021FANCYAD}, Question Answering \citep{ravichander-etal-2021-noiseqa}, Machine Reading Comprehension \citep{khashabi-etal-2018-looking}, Machine Translation (MT) \citep{king-falkedal-1990-using,isabelle-etal-2017-challenge}, and the more specific task of pronoun translation in MT \citep{guillou-hardmeier-2016-protest}.

The WMT 2021 Metrics shared task \citep{freitag-etal-2021-results} introduced the task of constructing contrastive challenge sets for the evaluation of MT metrics. Contrastive challenge sets aim to assess how well a given metric is able to discriminate between a \textit{good} and \textit{incorrect} translation of the \textit{source} text. The provision of a \textit{reference} translation allows for flexibility: it may be included for the assessment of reference-based (i.e. MT) metrics, or excluded for the assessment of reference-free (i.e. Quality Estimation (QE)) metrics.

We re-submitted \textsc{ACES}\footnote{The \textsc{ACES} dataset is available at \url{https://huggingface.co/datasets/nikitam/ACES}} \citep{amrhein-etal-2022-aces}, originally developed for the WMT 2022 challenge sets shared task \citep{freitag-etal-2022-results}, to the corresponding shared task at WMT 2023. \textsc{ACES} largely focuses on translation accuracy errors and consists of 68 phenomena ranging from simple perturbations at the word/character level to more complex errors based on discourse and real-world knowledge.
We report on both the performance of metrics submitted to WMT 2023, and on the incremental performance of those metrics that were submitted to both WMT 2022 and WMT 2023. We also repeat the analyses in \citet{amrhein-etal-2022-aces} for the WMT 2023 metrics to confirm whether the findings from WMT 2022 still hold.

\begin{figure*}
    \centering
    \includegraphics[width=\textwidth]{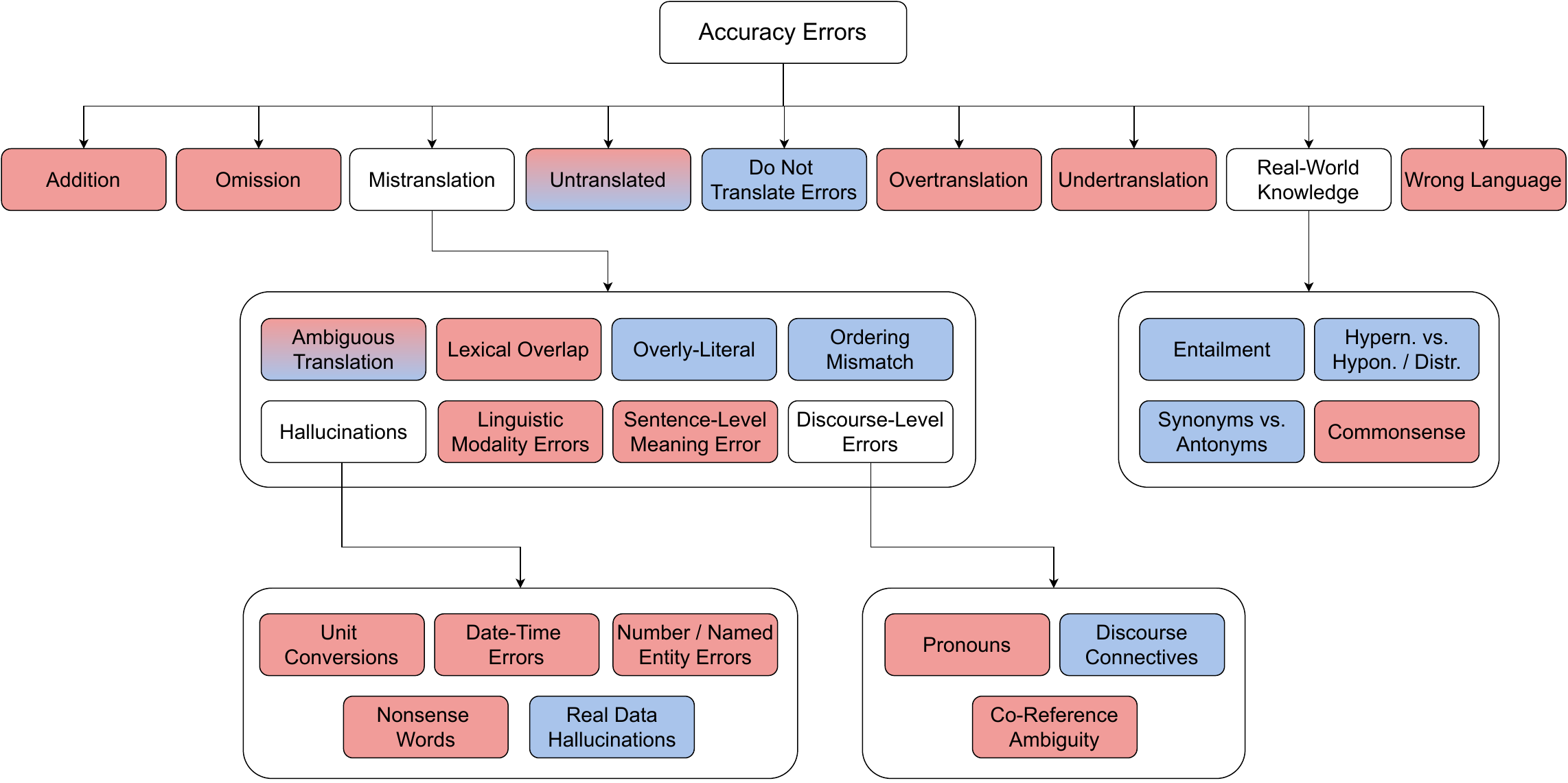}
    \caption{Diagram of the error categories on which our collection of challenge sets is based. Red means challenge sets are created automatically, and blue means challenge sets are created manually.}
    \label{fig:diagram}
\end{figure*}

Overall, we find similar trends to those observed last year. Again, we do not find one clear winner and whilst neural metrics tend to perform better than their non-neural counterparts, different categories of metrics exhibit different strengths and weaknesses. The major challenges identified in \citet{amrhein-etal-2022-aces} still hold: (i) reference-based metrics are still overly reliant on the reference and do not pay enough attention to the source, (ii) reference-based metrics still rely on surface-level overlap, and (iii) the over-reliance on multilingual embeddings still persists -- evidence from our analyses suggests that language agnostic representations present in the multilingual space may harm performance. Accordingly, our recommendations are also similar to those of last year. Metric developers should focus on: constructing ensembles of metrics with different design principles, developing metrics that also focus on information in the source, reducing dependence on surface-level overlap with the reference, and reassessing the impact of multilingual embeddings on MT evaluation. 

With respect to incremental performance changes between metrics submitted to both 2022 and 2023, we observe mixed results. Whilst improvements are observed for some metrics, there is a degradation in performance for other metrics. However, even for those metrics for which an overall improvement was observed, this improvement was inconsistent across the top-level categories in ACES. Further, the performance even degraded for some categories.


\section{ACES Overview}
\label{sec:aces_overview}
The Translation \textbf{A}ccuracy \textbf{C}halleng\textbf{E} Set (\textsc{ACES}) consists of 36,476 examples covering 146 language pairs and representing challenges from 68 linguistic phenomena. These phenomena are grouped into ten top-level categories: addition, omission, mistranslation, untranslated, do not translate errors, overtranslation, undertranslation, real-world knowledge, wrong language, and punctuation \footnote{Note that although the focus of ACES is on \textit{accuracy} errors, we also include a small set of \textit{fluency} errors for the punctuation category.}. The mistranslation and real-world knowledge categories are further sub-categorised to include additional fine-grained categories. We illustrate the broad accuracy error categories in Fig~\ref{fig:diagram} and give examples for each of the top-level categories in Appendix~\ref{sec:top_category}.

The focus of \textsc{ACES} is on translation accuracy errors, reflecting the need to evaluate contemporary MT systems that are capable of producing fluent but potentially error-prone output. The selection of the top-level categories in the \textsc{ACES} error hierarchy is based on the \textit{Accuracy} class in the Multidimensional Quality Metrics (MQM) ontology \citep{lommel2014}, and extended to include translations defying \textit{real-world knowledge} and translations in the \textit{wrong language}. \textsc{ACES} includes a wide range of phenomena ranging from simple perturbations that involve the omission/addition of characters or tokens, to more complex scenarios involving mistranslations e.g. ambiguity or hallucinations in translation, untranslated elements of a sentence, discourse-level phenomena, and real-world knowledge.

\begin{table*}[ht!]
    \small
    \begin{tabular}{rl}
        \toprule
        & \textbf{\textit{Mistranslation - Overly literal (Idioms)}} \\
        SRC (de): & Er hat versucht, mir die Spielregeln zu erklären, aber \textbf{ich verstand nur Bahnhof}. \\
        REF (en): & He tried to explain the rules of the game to me, but \textbf{I did not understand them}. \\
        \cmark: & He tried to explain the rules of the game to me, but \textbf{it was all Greek to me}. \\
        \xmark: & He tried to explain the rules of the game to me, but \textbf{I only understood train station}.\\
        \midrule
        & \textbf{\textit{Real-world Knowledge - Commonsense}} \\
        SRC (en): & Die Luft im Haus war kühler als in der Wohnung. \\
        REF (de): & The air in the house was cooler than in the apartment. \\
        \cmark: & The air in the house was cooler than in the apartment because \textbf{the apartment} had a broken air conditioner. \\
        \xmark: & The air in the house was cooler than in the apartment because \textbf{the house} had a broken air conditioner.\\
     \bottomrule
    \end{tabular}
    \caption{Examples from the Mistranslation and Real-world Knowledge categories in \textsc{ACES}. An example consists of a source sentence (SRC), reference (REF), good (\cmark) and incorrect (\xmark) translations, and a phenomenon label indicating the error type. en: English, de: German. Top: the German idiom ``ich verstand nur Bahnhof'' has been translated in an overly-literal way in the incorrect translation. Bottom: the incorrect translation contains an error in commonsense reasoning as to why the air in the house was cooler than in the apartment.}
    \label{tab:ACES_examples}
\end{table*}

Each \textsc{ACES} example consists of a \textit{source} sentence, a \textit{reference} translation, a \textit{phenomenon label} indicating the error type, and two
translation hypotheses: an \textit{incorrect} translation containing an error relating to the phenomenon of interest, and a \textit{good} translation. Several examples from \textsc{ACES} are presented in Table~\ref{tab:ACES_examples}. In the top example, from the \textit{Mistranslation} error category, the incorrect translation contains an \textit{overly literally} translation of the German \textit{idiom} ``ich verstand nur Bahnhof'' (corresponding to the English expression ``it was all Greek to me''). In the bottom example, from the \textit{Real-world Knowledge} error category, both the good and incorrect translations contain additional information not present in the source sentence, however, the incorrect translation contains an error in \textit{commonsense} reasoning as to why the air in the house was cooler than the apartment. 

\textsc{ACES} examples were constructed from pre-existing datasets, using a range of automatic, semi-automatic, and manual methods. A detailed description of each of the phenomena in \textsc{ACES} can be found in \citet{amrhein-etal-2022-aces}.

\section{Related Work}
\label{sec:related_work}
Challenge sets have been used for several tasks (\citet{li-etal-2017-bibi, mccoy2019non, ravichander-etal-2021-noiseqa}, \textit{inter alia}) to investigate the behaviour of these tasks under a specific phenomenon rather than the standard test distribution \citep{popovic-castilho-2019-challenge}. Lately, with the success of neural metrics, the development of challenge sets for MT evaluation has promoted great interest in studying the strengths and weaknesses of these metrics. We summarise here recent work on challenge sets for MT metric evaluation.

DEMETR \citep{karpinska-etal-2022-demetr}, which comprises 31K English examples translated from ten languages, was developed for evaluating MT metric sensitivity to a range of 35 different types of linguistic perturbations, belonging to semantic, syntactic, and morphological error categories. These were divided into minor, major, and critical errors according to the type of perturbation, similar to the grading of error categories to compute the weighted \textsc{ACES}-Score. As in \textsc{ACES}, example generation was carefully designed to form minimal pairs such that the perturbed translation only differs from the actual translation in one aspect. The application of DEMETR in evaluating a suite of baseline metrics revealed a similar pattern to the analyses in \citet{amrhein-etal-2022-aces} - that metric performance varies considerably across the different error categories, often with no clear \textit{winner}. It is worth noting that DEMETR and \textsc{ACES} each have their respective advantages: all examples in DEMETR have been verified by human annotators; \textsc{ACES} provides broader coverage in terms of both languages and linguistic phenomena.

In addition to \textsc{ACES}, three other datasets were submitted to the WMT 2022 challenge sets shared task \citep{freitag-etal-2022-results}: SMAUG 
\citep{alves-etal-2022-robust}, the HWTSC challenge set \citep{chen-etal-2022-exploring}, and the DFKI challenge set \citep{avramidis-macketanz-2022-linguistically}. These datasets differ from \textsc{ACES} in terms of their size, and the languages and phenomena/categories they cover (see Table~\ref{tab:wmt22_challenge_set_comparison}).

\begin{table}[]
    \centering
    \small
    \begin{tabular}{lcccc}
    \toprule
         & Ex. & Lang. & Phenomena & Categories \\
         & & pairs & & \\
    \midrule
        SMAUG & 632 & 2 & 5 & 5 \\
        HWTSC & 721 & 1 & 5 & 5 \\
        DFKI & 19,347 & 1 & >100 & 14 \\
        ACES & 36,476 & 146 & 68 & 10 \\
    \bottomrule
    \end{tabular}
    \caption{Comparison of challenge sets for MT metric evaluation in terms of: \textbf{Ex}amples, \textbf{Lang}uage\textbf{-pairs}, Phenomena, and Categories.}
    \label{tab:wmt22_challenge_set_comparison}
\end{table}

Both SMAUG and HWTSC contained five different phenomena each pertaining to a single category of critical error for meaning change. In comparison, the DFKI challenge set has over 100 linguistically motivated phenomena, organised into 14 categories. Whereas the aim of \textsc{ACES} was to provide a broad coverage of language pairs, the other datasets provide an in-depth focus on specific language pairs: SMAUG (pt$\leftrightarrow$en and es$\rightarrow$en), DFKI (de$\leftrightarrow$en), and HWTSC (zh$\leftrightarrow$en). Whilst there is a clear overlap between the \textsc{ACES} phenomena and those in SMAUG and HWTSC, many of the phenomena in the DFKI dataset are complementary such that in the case of evaluating metrics for the German-English pair, metric developers might consider benchmarking on both datasets.

\section{Metrics}
\label{sec:metrics}
We list below the metrics that participated in the 2023 challenge set shared task and the baselines provided by the organisers.

\subsection{Baseline Metrics}
\label{sec:metrics_baseline}

\noindent \textbf{BERTScore} \citep{DBLP:conf/iclr/ZhangKWWA20} uses contextual embeddings from pre-trained language models to compute the cosine similarity between the tokens in the hypothesis and the reference translation. The resulting similarity matrix is used to compute precision, recall, and F1-scores.\\

\noindent \textbf{BLEURT-20} \citep{sellam-etal-2020-learning} is a BERT-based \citep{devlin-etal-2019-bert} regression model, which is first trained on scores produced by automatic metrics/similarity of pairs of reference sentences and their corrupted counterparts. It is then fine-tuned on WMT human evaluation data to provide a score for a hypothesis given a reference translation.\\

\textbf{BLEU} \citep{papineni-etal-2002-bleu} compares the token-level n-grams in the hypothesis with those in the reference translation. It then computes a precision score weighted by a brevity penalty.\\

\noindent \textbf{chrF} \citep{popovic-2017-chrf} provides a character n-gram F-score by computing overlaps between the hypothesis and reference translation.\\

\noindent \textbf{COMET-22} \citep{rei-etal-2022-comet} is an ensemble between a vanilla \textsc{COMET} model \citep{rei-etal-2020-comet} trained with Direct Assessment (DA) scores and a multitask model that is trained on regression (MQM regression) and sequence tagging (OK/BAD word identification from MQM span annotations). These models are ensembled together using a hyperparameter search that weights different features extracted from these two evaluation models and combines them into a single score.
The vanilla \textsc{COMET} model is trained with DAs ranging from 2017 to 2020 while the Multitask model is trained using DAs ranging from 2017 to 2020 plus MQM annotations from 2020 (except for en-ru which uses TedTalk annotations from 2021).\\

\noindent \textbf{COMET-Kiwi} \citep{rei-etal-2022-comet} ensembles two QE models similarly to \textsc{COMET-22}. The first model follows the classic Predictor-Estimator QE architecture where MT and source are encoded together. This model is trained on DAs ranging from 2017 to 2019 and then fine-tuned on DAs from MLQE-PE (the official DA from the QE shared task). The second model is the same multitask model used in the \textsc{COMET-22} submission but without access to a reference translation. This means that this model is a multitask model trained on regression and sequence tagging. Both models are ensembled together using a hyperparameter search that weights different features extracted from these two QE models and combines them into a single score.\\

\noindent \textbf{f200spBLEU} \citep{goyal-etal-2022-flores} computes BLEU over text tokenised with a single language-agnostic SentencePiece subword model. For the f200spBLEU version of spBLEU used in this year's shared task, the SentencePiece tokeniser \citep{kudo-richardson-2018-sentencepiece} was trained using data from the FLORES-200 languages.\\

\noindent \textbf{MS-COMET-22} \citep{kocmi-etal-2022-ms} is built on top of the COMET \citep{rei-etal-2020-comet} architecture. It is trained on a set of human judgments several times larger -- covering 113 languages and 15 domains. Furthermore, the authors propose filtering out those human judgements with potentially low quality. \textsc{MS-COMET-22} is a reference-based metric that receives the source, the MT hypothesis and the human reference as input.\\

\noindent \textbf{Random-sysname} is a random baseline. The metric takes the name of the system as the only parameter. It uses a discrete score. Segment-level scores follow a Gaussian distribution around mean value X (in the range 0-9) and a standard deviation of 2. The mean X is calculated from the name of the system as: $X = sha256(sysname)[0] \% 10$

The idea behind this baseline is two-fold. Firstly, having a baseline showing how a random metric would perform could help to put scores into context (in particular, pairwise accuracy can create a perception of great performance while 50\% is just a toss of a coin). Secondly, it could help to detect errors in metric meta-evaluations.\\

\noindent \textbf{YiSi-1} \citep{lo-2019-yisi}
measures the semantic similarity between the hypothesis and the reference translation by using cosine similarity scores of multilingual representations at the lexical level. It optionally uses a semantic role labeller to obtain structural similarity. Finally, a weighted F-score based on structural and lexical similarity is used for scoring the hypothesis against the reference translation.

\subsection{Metrics Submitted to WMT 2023}
\label{sec:metrics_participant}
We list the descriptions of the metrics submitted to WMT 2023 by the metric developers and refer the reader to the relevant system description papers for further details.\\

\noindent \textbf{Embed\_Llama} relies on pretrained Llama 2 embeddings, without any finetuning, to transform sentences into a vector space that establishes connections between geometric and semantic proximities. This metric draws inspiration from Word2vec and utilizes cosine distance for the purpose of estimating similarity or dissimilarity between sentences. \\


\noindent \textbf{MetricX-23 and MetricX-23-QE} are learned reference-based and reference-free (respectively) regression metrics based on the mT5 encoder-decoder language model. They further finetune the mT5-XXL checkpoint on direct assessment data from 2015-2020 and MQM data from 2020 to 2021 as well as synthetic data.\\

\noindent \textbf{Tokengram\_F} is an F-score-based evaluation metric that is heavily inspired by chrF++. By replacing word-grams with token-grams obtained from contemporary tokenization algorithms, tokengram\_F captures similarities between words sharing the same semantic roots and thus obtains more accurate ratings. \\

\noindent \textbf{Partokengram\_F} we did not receive a description of this metric.\\

\noindent \textbf{XCOMET} is a new COMET-base model that is trained to identify errors in sentences along with a final quality score and thus leads to an explainable neural metric. The metric is optimised towards regression and error span detection simultaneously. The same model may be used both with references (\textsc{XCOMET}) and without references (\textsc{XCOMET-QE}). The models are built using XLM-R XL and XXL, thus \textsc{XCOMET-XL} has 3.5B parameters and \textsc{XCOMET-XXL} has 10.7B parameters. The metric is trained in stages where it first sees DAs and then is fine-tuned with MQM. \textsc{XCOMET-Ensemble} is an ensemble between 1 XL and 2 XXL checkpoints that result from these training stages.\\

\noindent \textbf{XLsim} is a supervised reference-based metric that regresses on human scores provided by WMT (2017-2022). Using a cross-lingual language model (XLM-RoBERTa \citep{conneau-etal-2020-unsupervised}), a supervised model is trained using a Siamese network architecture with CosineSimilarityLoss. \textbf{\textsc{XLsimQE}} is the reference-free variant of this metric. \\

\noindent \textbf{Cometoid22} is a reference-free metric created using knowledge distillation from reference-based metrics. First, using COMET-22 as a teacher metric, the MT outputs submitted to the WMT News/General Translation task since 2009 are scored. Next, a student metric, called Cometoid22, is trained to mimic the teacher scores without using reference translation. The student metric has the same architecture as COMET-QE, and is initialised with pre-trained weights from the multilingual language model InfoXLM. Three variants were submitted: cometoid22-wmt{21,22,23}, where the suffix indicates the training data cut-off year.\\

\noindent \textbf{COMETKiwi-XL and COMETKiwi-XXL} use the same COMETKiwi model architecture from WMT 2022 but replace InfoXLM with XLM-R XL and XXL (for \textsc{COMETKiwi-XL} and \textsc{COMETKiwi-XXL} respectively).\\

\noindent \textbf{KG-BERTScore} incorporates a multilingual knowledge graph into BERTScore and generates the final evaluation score by linearly combining the results of KGScore and BERTScore. In contrast to last year, COMET-QE is used to calculate BERTScore.\\


\noindent \textbf{GEMBA-MQM} is an LLM-enabled metric for error quality span marking. It uses three-shot prompting with a GPT-4 model. In contrast to EAPrompt \citep{lu2023error}, it does not require language-specific examples and requires only a single prompt.\\


\section{Results}
\label{sec:results}

\subsection{Phenomena-level Results}
\label{subsec:phenomena_results}
As in \citet{amrhein-etal-2022-aces} we begin by providing a broad overview of metric performance on the different phenomena categories, before conducting more detailed analyses (see Section~\ref{subsec:results_analysis}). We restrict the overview to the metrics which provide a) segment-level scores and b) scores for all language pairs and directions in \textsc{ACES}. Out of the metrics that participated, 33 fulfil these criteria: 10 baselines, 11 reference-based, and 12 reference-free metrics.

We first compute the Kendall’s tau-like correlation scores\footnote{Evaluation scripts are available here: \url{https://github.com/EdinburghNLP/ACES}} \citep{freitag-etal-2021-results,freitag-etal-2022-results} for all of the \textsc{ACES} examples. This metric measures the number of times a metric scores the good translation above the incorrect translation (concordant) and equal to or lower than the incorrect translation (discordant): 

\begin{center}
\small
\begin{align*}
    \tau = \frac{concordant - discordant}{concordant + discordant}
\end{align*}
\end{center}
\vspace{0.25cm}

We then report the average score over all examples belonging to each of the nine top-level accuracy categories in \textsc{ACES}, plus the fluency category \textit{punctuation} (see Table~\ref{tab:analysis_overview}). 
In addition, we compute the \textsc{ACES}-Score, a weighted combination of the top-level categories, which allows us to identify high-level performance trends of the metrics (see Equation~\ref{eq:aces-score}). Note that the \textsc{ACES}-Score ranges from -29.1 (all phenomena have a correlation of -1) to 29.1 (all phenomena have a correlation of +1).

\begin{equation}
\small
\label{eq:aces-score}
\textsc{ACES} = sum \left\{
\begin{aligned}
    \quad 5 * \tau_{\text{addition}}\\
   \quad  5 * \tau_{\text{omission}}\\
   \quad  5 * \tau_{\text{mistranslation}}\\
   \quad  1 * \tau_{\text{untranslated}}\\
   \quad  1 * \tau_{\text{do not translate}}\\
   \quad  5 * \tau_{\text{overtranslation}}\\
   \quad  5 * \tau_{\text{undertranslation}}\\
   \quad  1 * \tau_{\text{real-world knowledge}}\\
   \quad  1 * \tau_{\text{wrong language}}\\
   \quad  0.1 * \tau_{\text{punctuation}}\\
\end{aligned}
\right\} \vspace{0.25cm}
\end{equation}

Overall, the best-performing metrics submitted to this year's shared task, according to the \textsc{ACES}-Score, are \textsc{COMETKiwi} (a reference-free baseline metric), and \textsc{KG-BERTScore} (a reference-free metric). BLEU remains one of the worst-performing metrics, with only the random baseline, \textsc{Random-sysname}, achieving a lower \textsc{ACES}-Score. \textsc{XCOMET-Ensemble} is the top ranking among the reference-based metrics. We caution that we developed ACES to investigate strengths and weaknesses of metrics on a phenomena level -- hence, we advise the reader not to draw any conclusions based solely on the \textsc{ACES}-Score.

As observed in \citet{amrhein-etal-2022-aces} the performance of the metrics is highly variable, with no clear \textit{winner} in terms of performance across all of the top-level \textsc{ACES} categories. We also observe similar trends in terms of the most challenging categories (\textit{addition}, \textit{undertranslation}, \textit{real-world knowledge}, and \textit{wrong language}).  We find that, unlike last year, some metrics perform similarly to or worse than the baseline metrics. In particular, \textsc{embed\_llama} and \textsc{GEMBA-MQM} which are designed using Large Language Models (LLMs), struggle with this challenge set. This suggests that we need better design strategies in using the rich representations from LLMs for MT evaluation. In general, we find that reference-free metrics perform on par or better than reference-based metrics.

In terms of performance across the top-level categories, we also observe variation in the performance of metrics belonging to the baseline, reference-based, and reference-free groups. The reference-free group exhibits overall stronger performance compared with the other groups, but in particular for the \textit{mistranslation}, \textit{overtranslation}, \textit{undertranslation}, and \textit{real-world knowledge} categories. 

\begin{sidewaystable*}[ht]
\small
\setlength{\tabcolsep}{3.75pt}
\centering
\begin{tabular}{@{}lccccccccccc@{}}
\toprule
         & addition & omission & mistranslation & untranslated & do not & overtranslation & undertranslation & real-world & wrong & punctuation & ACES-Score \\
         &  &  &  &  & translate &  &  & knowledge & language &  & \\
\midrule
\textit{Examples}           & \textit{999}                          & \textit{999}                          & \textit{24457}                              & \textit{1300}                             & \textit{100}                                  & \textit{1000}                                & \textit{1000}                                 & \textit{2948}                                     & \textit{2000}                               & \textit{1673}                            &            \\
\midrule
BERTscore          & \colorbox[HTML]{B2EAB1}{\textbf{\phantom{-}0.872}}                        & \phantom{-}0.754                        & \phantom{-}0.318                              & \phantom{-}0.771                            & \phantom{-}0.940                                & -0.186                              & -0.288                               & \phantom{-}0.030                                    & \phantom{-}0.551                              & \colorbox[HTML]{B2EAB1}{\textbf{\phantom{-}0.844}}                           & \phantom{1}9.722                          \\
BLEU               & \phantom{-}0.742                        & \phantom{-}0.427                        & -0.227                             & \phantom{-}0.353                            & \phantom{-}0.580                                & -0.838                              & -0.856                               & -0.768                                   & \phantom{-}0.660                              & \phantom{-}0.704                           & -2.862                         \\
BLEURT-20          & \phantom{-}0.435                        & \phantom{-}0.812                        & \phantom{-}0.427                              & \phantom{-}0.743                            & \phantom{-}0.860                                & \phantom{-}0.202                               & \phantom{-}0.014                                & \phantom{-}0.388                                    & \phantom{-}0.536                              & \phantom{-}0.708                           & 12.048                         \\
chrF               & \phantom{-}0.644                        & \phantom{-}0.784                        & \phantom{-}0.162                              & \colorbox[HTML]{B2EAB1}{\textbf{\phantom{-}0.781}}                            & \colorbox[HTML]{B2EAB1}{\textbf{\phantom{-}0.960}}                                & -0.696                              & -0.592                               & -0.294                                   & \phantom{-}0.693                              & \phantom{-}0.773                           & \phantom{-}3.728                          \\
COMET-22           & \phantom{-}0.295                        & \phantom{-}0.822                        & \phantom{-}0.402                              & \phantom{-}0.718                            & \phantom{-}0.820                                & \phantom{-}0.502                               & \phantom{-}0.258                                & \phantom{-}0.382                                    & \phantom{-}0.078                              & \phantom{-}0.673                           & 13.458                         \\
CometKiwi          & \phantom{-}0.536                        & \colorbox[HTML]{B2EAB1}{\textbf{\phantom{-}0.918}}                        & \phantom{-}0.614                              & -0.105                           & \phantom{-}0.520                                & \phantom{-}0.766                               & \phantom{-}0.604                                & \phantom{-}0.577                                    & -0.307                             & \phantom{-}0.765                           & \colorbox[HTML]{B2EAB1}{\textbf{17.951}}                         \\
f200spBLEU         & \phantom{-}0.666                        & \phantom{-}0.584                        & -0.082                             & \phantom{-}0.680                            & \phantom{-}0.920                                & -0.752                              & -0.794                               & -0.394                                   & \phantom{-}0.657                              & \phantom{-}0.708                           & \phantom{1}0.041                          \\
MS-COMET-QE-22     & -0.179                       & \phantom{-}0.674                        & \phantom{-}0.440                              & \phantom{-}0.394                            & \phantom{-}0.300                                & \phantom{-}0.524                               & \phantom{-}0.382                                & \phantom{-}0.262                                    & -0.195                             & \phantom{-}0.632                           & 10.027                         \\
Random-sysname     & -0.117                       & -0.117                       & -0.116                             & -0.083                           & -0.100                               & -0.118                              & -0.152                               & -0.245                                   & -0.113                             & -0.074                          & -3.648                         \\
YiSi-1             & \phantom{-}0.766                        & \phantom{-}0.868                        & \phantom{-}0.354                              & \phantom{-}0.720                            & \phantom{-}0.940                                & -0.062                              & -0.076                               & \phantom{-}0.110                                    & \phantom{-}0.421                              & \phantom{-}0.763                           & 11.517                         \\
\midrule
eBLEU              & \phantom{-}0.674                        & \phantom{-}0.682                        & \phantom{-}0.197                              & \phantom{-}0.739                            & \phantom{-}0.880                                & -0.662                              & -0.684                               & -0.042                                   & \colorbox[HTML]{B2EAB1}{\textbf{\phantom{-}0.771}}                              & \phantom{-}0.270                           & \phantom{1}3.406                          \\
embed\_llama       & \phantom{-}0.211                        & \phantom{-}0.457                        & \phantom{-}0.016                              & \phantom{-}0.503                            & \phantom{-}0.400                                & -0.170                              & -0.492                               & -0.165                                   & \phantom{-}0.154                              & \phantom{-}0.476                           & \phantom{1}1.054                          \\
MetricX-23         & -0.027                       & \phantom{-}0.568                        & \phantom{-}0.578                              & \phantom{-}0.473                            & \phantom{-}0.800                                & \phantom{-}0.790                               & \phantom{-}0.586                                & \phantom{-}0.766                                    & -0.486                             & \phantom{-}0.636                           & 14.091                         \\
MetricX-23-b       & -0.135                       & \phantom{-}0.622                        & \phantom{-}0.572                              & \phantom{-}0.613                            & \phantom{-}0.860                                & \phantom{-}0.772                               & \phantom{-}0.568                                & \phantom{-}0.749                                    & -0.444                             & \phantom{-}0.532                           & 13.826                         \\
MetricX-23-c       & -0.015                       & \phantom{-}0.794                        & \phantom{-}0.617                              & \phantom{-}0.611                            & \phantom{-}0.800                                & \phantom{-}0.740                               & \phantom{-}0.526                                & \colorbox[HTML]{B2EAB1}{\textbf{\phantom{-}0.783}}                                    & -0.629                             & \phantom{-}0.527                           & 14.929                         \\
partokengram\_F    & \phantom{-}0.087                        & \phantom{-}0.191                        & -0.034                             & \phantom{-}0.310                            & \phantom{-}0.140                                & -0.042                              & -0.028                               & \phantom{-}0.032                                    & \phantom{-}0.508                              & \phantom{-}0.171                           & \phantom{1}1.878                          \\
tokengram\_F       & \phantom{-}0.698                        & \phantom{-}0.758                        & \phantom{-}0.160                              & \phantom{-}0.779                            & \colorbox[HTML]{B2EAB1}{\textbf{\phantom{-}0.960}}                                & -0.732                              & -0.632                               & -0.273                                   & \phantom{-}0.687                              & \phantom{-}0.830                           & \phantom{1}3.492                          \\
XCOMET-Ensemble    & \phantom{-}0.311                        & \phantom{-}0.786                        & \phantom{-}0.663                              & \phantom{-}0.379                            & \phantom{-}0.780                                & \colorbox[HTML]{B2EAB1}{\textbf{\phantom{-}0.794}}                               & \phantom{-}0.612                                & \phantom{-}0.708                                    & -0.423                             & \phantom{-}0.595                           & 17.336                         \\
XCOMET-XL          & \phantom{-}0.169                        & \phantom{-}0.542                        & \phantom{-}0.570                              & \phantom{-}0.222                            & \phantom{-}0.800                                & \phantom{-}0.656                               & \phantom{-}0.464                                & \phantom{-}0.582                                    & -0.367                             & \phantom{-}0.220                           & 13.264                         \\
XCOMET-XXL         & -0.119                       & \phantom{-}0.413                        & \phantom{-}0.547                              & \phantom{-}0.234                            & \phantom{-}0.600                                & \phantom{-}0.736                               & \phantom{-}0.568                                & \phantom{-}0.508                                    & -0.507                             & \phantom{-}0.509                           & 11.610                         \\
XLsim              & \phantom{-}0.429                        & \phantom{-}0.618                        & \phantom{-}0.153                              & \phantom{-}0.643                            & \phantom{-}0.820                                & -0.210                              & -0.290                               & -0.044                                   & \phantom{-}0.392                              & \phantom{-}0.753                           & \phantom{1}5.386                          \\
\midrule
cometoid22-wmt21   & -0.339                       & \phantom{-}0.658                        & \phantom{-}0.493                              & -0.076                           & \phantom{-}0.280                                & \phantom{-}0.670                               & \phantom{-}0.566                                & \phantom{-}0.362                                    & -0.454                             & \phantom{-}0.608                           & 10.409                         \\
cometoid22-wmt22   & -0.301                       & \phantom{-}0.674                        & \phantom{-}0.493                              & -0.119                           & \phantom{-}0.280                                & \phantom{-}0.686                               & \phantom{-}0.538                                & \phantom{-}0.340                                    & -0.472                             & \phantom{-}0.599                           & 10.534                         \\
cometoid22-wmt23   & -0.253                       & \phantom{-}0.702                        & \phantom{-}0.502                              & -0.046                           & \phantom{-}0.420                                & \phantom{-}0.750                               & \phantom{-}0.590                                & \phantom{-}0.362                                    & -0.319                             & \phantom{-}0.557                           & 11.926                         \\
CometKiwi-XL       & \phantom{-}0.239                        & \phantom{-}0.828                        & \phantom{-}0.624                              & \phantom{-}0.239                            & \phantom{-}0.440                                & \phantom{-}0.762                               & \phantom{-}0.560                                & \phantom{-}0.563                                    & -0.380                             & \phantom{-}0.630                           & 15.988                         \\
CometKiwi-XXL      & \phantom{-}0.361                        & \phantom{-}0.828                        & \phantom{-}0.653                              & \phantom{-}0.414                            & \phantom{-}0.320                                & \phantom{-}0.774                               & \phantom{-}0.560                                & \phantom{-}0.683                                    & -0.537                             & \phantom{-}0.503                           & 16.809                         \\
GEMBA-MQM          & \phantom{-}0.037                        & \phantom{-}0.281                        & \phantom{-}0.153                              & \phantom{-}0.094                            & \phantom{-}0.140                                & \phantom{-}0.466                               & \phantom{-}0.276                                & \phantom{-}0.268                                    & -0.150                             & \phantom{-}0.015                           & \phantom{1}6.419                          \\
KG-BERTScore       & \phantom{-}0.538                        & \phantom{-}0.912                        & \phantom{-}0.585                              & -0.206                           & \phantom{-}0.700                                & \phantom{-}0.772                               & \phantom{-}0.606                                & \phantom{-}0.594                                    & -0.307                             & \phantom{-}0.654                           & 17.906                         \\
MetricX-23-QE      & \phantom{-}0.045                        & \phantom{-}0.678                        & \phantom{-}0.654                              & \phantom{-}0.379                            & \phantom{-}0.460                                & \phantom{-}0.772                               & \phantom{-}0.612                                & \phantom{-}0.654                                    & -0.702                             & \phantom{-}0.226                           & 14.614                         \\
MetricX-23-QE-b    & \phantom{-}0.027                        & \phantom{-}0.760                        & \phantom{-}0.663                              & \phantom{-}0.489                            & \phantom{-}0.480                                & \phantom{-}0.758                               & \colorbox[HTML]{B2EAB1}{\textbf{\phantom{-}0.620}}                                & \phantom{-}0.647                                    & -0.673                             & \phantom{-}0.256                           & 15.106                         \\
MetricX-23-QE-c    & -0.115                       & \phantom{-}0.664                        & \colorbox[HTML]{B2EAB1}{\textbf{\phantom{-}0.721}}                              & \phantom{-}0.384                            & \phantom{-}0.340                                & \phantom{-}0.726                               & \phantom{-}0.618                                & \phantom{-}0.753                                    & -0.712                             & \phantom{-}0.375                           & 13.873                         \\
XCOMET-QE-Ensemble & \phantom{-}0.277                        & \phantom{-}0.754                        & \phantom{-}0.644                              & \phantom{-}0.181                            & \phantom{-}0.720                                & \phantom{-}0.764                               & \phantom{-}0.582                                & \phantom{-}0.626                                    & -0.519                             & \phantom{-}0.449                           & 16.156                         \\
XLsimQE            & \phantom{-}0.205                        & \phantom{-}0.383                        & \phantom{-}0.087                              & -0.694                           & \phantom{-}0.940                                & \phantom{-}0.454                               & \phantom{-}0.352                                & \phantom{-}0.042                                    & \phantom{-}0.307                              & \phantom{-}0.671                           & \phantom{1}8.070                         \\
\midrule
Average            & \phantom{-}0.232                        & \phantom{-}0.639                        & \phantom{-}0.382                              & \phantom{-}0.349                            & \phantom{-}0.609                                & \phantom{-}0.314                               & \phantom{-}0.187                                & \phantom{-}0.289                                    & -0.069                             & \phantom{-}0.532                           & 10.002                        
\\
\bottomrule
\end{tabular}
\caption{Average Kendall’s tau-like correlation results for the \textsc{ACES} top-level categories and \textsc{ACES}-Scores (final column). Metrics are grouped into baseline (top), and participating reference-based (middle) and reference-free (bottom) metrics. Note that \textit{Average} is an average over averages. Best results are highlighted in green.}
\label{tab:analysis_overview}
\end{sidewaystable*}
\clearpage



\subsection{Mistranslation Results}
Next, we drill down to the fine-grained categories of the largest ACES category: \textit{mistranslation}. We present metric performance for the sub-level categories (\textit{discourse}, \textit{hallucination}, and \textit{other}) in Table~\ref{tab:analysis_mistranslation}. The \textit{discourse} sub-category includes errors involving the mistranslation of discourse-level phenomena such as pronouns and discourse connectives. \textit{Hallucination} includes errors at the word level that could occur due to hallucination by an MT model, for example, the use of wrong units, dates, times, numbers or named entities, as well as hallucinations at the subword level that result in nonsensical words. The \textit{other} cub-category covers all other categories of mistranslation errors including overly-literal translations (see example in Table~\ref{tab:ACES_examples}) and the introduction of ambiguities in the translation output. Again, as in 2022, we find that performance on the different sub-categories is highly variable, with no clear \textit{winner} among the metrics. We also make similar observations to those in \citet{amrhein-etal-2022-aces}, that the hallucination phenomena are generally more challenging than discourse-level phenomena; performance on the hallucination sub-category is poor overall. 

\begin{table}[h!]
\centering
\small
\setlength{\tabcolsep}{5pt}
\setlength{\fboxsep}{0.5pt} 
\begin{tabular}{@{}lccc@{}}
\toprule
                    & \textbf{disco.} & \textbf{halluci.} & \textbf{other}         \\ 
\midrule
\textbf{\textit{Examples}}           & \textbf{\textit{3698}}                                           & \textbf{\textit{10270}}                                              & \textbf{\textit{10489}}                                      \\
\midrule
BERTscore          & \phantom{-}0.563                                          & -0.062                                             & \phantom{-}0.361                                      \\
BLEU               & -0.042                                         & -0.418                                             & -0.250                                     \\
BLEURT-20          & \phantom{-}0.695                                          & \phantom{-}0.141                                              & \phantom{-}0.398                                      \\
chrF               & \phantom{-}0.406                                          & -0.138                                             & \phantom{-}0.160                                      \\
COMET-22           & \phantom{-}0.657                                          & \phantom{-}0.113                                              & \phantom{-}0.383                                      \\
CometKiwi          & \phantom{-}0.779                                          & \phantom{-}0.465                                              & \phantom{-}0.580                                      \\
f200spBLEU         & \phantom{-}0.095                                          & -0.190                                             & -0.150                                     \\
MS-COMET-QE-22     & \phantom{-}0.631                                          & \phantom{-}0.240                                              & \phantom{-}0.417                                      \\
Random-sysname     & -0.117                                         & -0.122                                             & -0.111                                     \\
YiSi-1             & \phantom{-}0.608                                          & \phantom{-}0.017                                              & \phantom{-}0.366                                      \\
\midrule
eBLEU              & \phantom{-}0.374                                          & -0.166                                             & \phantom{-}0.282                                      \\
embed\_llama       & -0.089                                         & -0.140                                             & \phantom{-}0.189                                      \\
MetricX-23         & \phantom{-}0.757                                          & \phantom{-}0.663                                              & \phantom{-}0.393                                      \\
MetricX-23-b       & \phantom{-}0.749                                          & \phantom{-}0.656                                              & \phantom{-}0.390                                      \\
MetricX-23-c       & \phantom{-}0.694                                          & \colorbox[HTML]{B2EAB1}{\textbf{\phantom{-}0.755}}                                              & \phantom{-}0.477                                      \\
partokengram\_F    & -0.062                                         & -0.101                                             & 0.027                                      \\
tokengram\_F       & \phantom{-}0.396                                          & -0.132                                             & \phantom{-}0.157                                      \\
XCOMET-Ensemble    & \colorbox[HTML]{B2EAB1}{\textbf{\phantom{-}0.791}}                                          & \phantom{-}0.566                                              & \phantom{-}0.626                                      \\
XCOMET-XL          & \phantom{-}0.706                                          & \phantom{-}0.482                                              & \phantom{-}0.521                                      \\
XCOMET-XXL         & \phantom{-}0.609                                          & \phantom{-}0.540                                              & \phantom{-}0.504                                      \\
XLsim              & \phantom{-}0.217                                          & -0.066                                             & \phantom{-}0.236                                      \\
\midrule
cometoid22-wmt21   & \phantom{-}0.782                                          & \phantom{-}0.286                                              & \phantom{-}0.400                                      \\
cometoid22-wmt22   & \phantom{-}0.748                                          & \phantom{-}0.290                                              & \phantom{-}0.423                                      \\
cometoid22-wmt23   & \phantom{-}0.758                                          & \phantom{-}0.223                                              & \phantom{-}0.478                                      \\
CometKiwi-XL       & \phantom{-}0.752                                          & \phantom{-}0.501                                              & \phantom{-}0.602                                      \\
CometKiwi-XXL      & \phantom{-}0.735                                          & \phantom{-}0.535                                              & \phantom{-}0.661                                      \\
GEMBA-MQM          & \phantom{-}0.076                                          & \phantom{-}0.291                                              & \phantom{-}0.127                                      \\
KG-BERTScore       & \phantom{-}0.685                                          & \phantom{-}0.466                                              & \phantom{-}0.580                                      \\
MetricX-23-QE      & \phantom{-}0.728                                          & \phantom{-}0.604                                              & \phantom{-}0.628                                      \\
MetricX-23-QE-b    & \phantom{-}0.694                                          & \phantom{-}0.617                                              & \phantom{-}0.666                                      \\
MetricX-23-QE-c    & \phantom{-}0.747                                          & \phantom{-}0.659                                              & \colorbox[HTML]{B2EAB1}{\textbf{\phantom{-}0.739}}                                      \\
XCOMET-QE-Ensemble & \phantom{-}0.702                                          & \phantom{-}0.558                                              & \phantom{-}0.651                                      \\
XLsimQE            & \phantom{-}0.053                                          & \phantom{-}0.050                                              & \phantom{-}0.134                                      \\
\midrule
Average            & \phantom{-}0.511                                          & \phantom{-}0.248                                              & \phantom{-}0.365     \\
\bottomrule
\end{tabular}
\caption{Average Kendall’s tau-like correlation results for the sub-level categories in mistranslation: \textbf{disco}urse-level, \textbf{halluci}nation, and \textbf{other} errors. Metrics are grouped into baseline (top), and participating reference-based (middle) and reference-free (bottom) metrics. Note that \textit{Average} is an average over averages. Best results are highlighted in green.}
\label{tab:analysis_mistranslation}
\end{table}


\subsection{Analysis}
\label{subsec:results_analysis}


We repeat the analyses we performed in \citet{amrhein-etal-2022-aces} for the metrics submitted to WMT 2023 to confirm whether our findings from WMT 2022 still hold. We highlight similar observations to those from WMT 2022 and summarise our insights below.

\subsubsection{How sensitive are metrics to the source?}
\label{analysis:source}
\begin{table*}[h!]
    \centering
    \small
    \resizebox{\linewidth}{!}{%

    \begin{tabular}{lcccccccc}
    \toprule
    & \multicolumn{2}{c}{\textbf{since}} & \multicolumn{2}{c}{\textbf{female}}  & \multicolumn{2}{c}{\textbf{male}} &\\
    \cmidrule(lr){2-3} \cmidrule(lr){4-5} \cmidrule(lr){6-7} \\
    & \textbf{causal} & \textbf{temp.} & \textbf{anti.} & \textbf{pro.} & \textbf{anti.} & \textbf{pro.} & \textbf{AVG}\\
    \cmidrule(lr){2-2} \cmidrule(lr){3-3} \cmidrule(lr){4-4} \cmidrule(lr){5-5} \cmidrule(lr){6-6}  \cmidrule(lr){7-7} \cmidrule(lr){8-8}  \\
    \textit{\textbf{Examples}} & \textit{106} & \textit{106} & \textit{1000} & \textit{806} & \textit{806} & \textit{1000} & \textit{3824}\\
    \midrule
Random-sysname & -0.075 & -0.019 & -0.146 & -0.156 & -0.109 & -0.154 & -0.110\\ \midrule
COMET-22 & -0.868 & \phantom{-}0.887 & -0.254 & \phantom{-}0.591 & -0.467 & \phantom{-}0.432 & \phantom{-}0.053\\
MetricX-23 & -1.000 & \colorbox[HTML]{B2EAB1}{\textbf{1.000}} & -0.864 & -0.062 & \phantom{-}0.062 & \phantom{-}0.870 & \phantom{-}0.001\\
MetricX-23-b & -1.000 & \colorbox[HTML]{B2EAB1}{\textbf{1.000}} & -0.790 & \phantom{-}0.112 & -0.092 & \phantom{-}0.780 & \phantom{-}0.002\\
MetricX-23-c & -0.849 & \phantom{-}0.849 & -0.998 & -0.581 & \colorbox[HTML]{B2EAB1}{\textbf{\phantom{-}0.576}} & \colorbox[HTML]{B2EAB1}{\textbf{\phantom{-}0.996}} & -0.001\\
XCOMET-Ensemble & -0.585 & \phantom{-}0.981 & \colorbox[HTML]{B2EAB1}{\textbf{\phantom{-}0.852}} & \phantom{-}0.948 & \phantom{-}0.273 & \phantom{-}0.922 & \colorbox[HTML]{B2EAB1}{\textbf{\phantom{-}0.565}}\\
XCOMET-XL & -0.698 & \phantom{-}0.906 & \phantom{-}0.456 & \colorbox[HTML]{B2EAB1}{\textbf{\phantom{-}0.960}} & -0.330 & \phantom{-}0.698 & \phantom{-}0.332\\
XCOMET-XXL & -0.868 & \phantom{-}0.925 & \phantom{-}0.372 & \phantom{-}0.675 & \phantom{-}0.541 & \phantom{-}0.918 & \phantom{-}0.427\\ \midrule
cometoid22-wmt21 & -0.698 & \phantom{-}0.868 & \phantom{-}0.580 & \phantom{-}0.950 & -0.787 & \phantom{-}0.022 & \phantom{-}0.156\\
cometoid22-wmt22 & -0.623 & \phantom{-}0.868 & \phantom{-}0.456 & \phantom{-}0.851 & -0.444 & \phantom{-}0.442 & \phantom{-}0.258\\
cometoid22-wmt23 & -0.566 & \phantom{-}0.925 & \phantom{-}0.342 & \phantom{-}0.851 & \phantom{-}0.117 & \phantom{-}0.844 & \phantom{-}0.419\\
CometKiwi & \phantom{-}0.075 & \colorbox[HTML]{B2EAB1}{\textbf{1.000}} & \colorbox[HTML]{B2EAB1}{\textbf{\phantom{-}0.990}} & \colorbox[HTML]{B2EAB1}{\textbf{\phantom{-}0.998}} & -0.171 & \phantom{-}0.440 & \phantom{-}0.555\\
CometKiwi-XL & \phantom{-}0.075 & \phantom{-}0.925 & \phantom{-}0.952 & \phantom{-}0.990 & \phantom{-}0.380 & \phantom{-}0.892 & \colorbox[HTML]{B2EAB1}{\textbf{\phantom{-}0.702}}\\
CometKiwi-XXL & \colorbox[HTML]{B2EAB1}{\textbf{\phantom{-}0.132}} & \phantom{-}0.943 & \phantom{-}0.932 & \phantom{-}0.995 & \phantom{-}0.241 & \phantom{-}0.796 & \phantom{-}0.673\\
GEMBA-MQM & -0.604 & \phantom{-}0.736 & \phantom{-}0.722 & \phantom{-}0.320 & -0.762 & -0.692 & -0.047\\
KG-BERTScore & \phantom{-}0.075 & \colorbox[HTML]{B2EAB1}{\textbf{1.000}} & \colorbox[HTML]{B2EAB1}{\textbf{\phantom{-}0.990}} & \colorbox[HTML]{B2EAB1}{\textbf{\phantom{-}0.998}} & -0.171 & \phantom{-}0.440 & \phantom{-}0.555\\
MS-COMET-QE-22 & -0.283 & \phantom{-}0.811 & -0.194 & \phantom{-}0.323 & \phantom{-}0.243 & \phantom{-}0.692 & \phantom{-}0.265\\
MetricX-23-QE & -0.472 & \phantom{-}0.736 & \phantom{-}0.974 & \phantom{-}0.995 & \phantom{-}0.117 & \phantom{-}0.816 & \phantom{-}0.528\\
MetricX-23-QE-b & -0.566 & \phantom{-}0.868 & \phantom{-}0.968 & \phantom{-}0.995 & \phantom{-}0.722 & \colorbox[HTML]{B2EAB1}{\textbf{\phantom{-}0.968}} & \phantom{-}0.659\\
MetricX-23-QE-c & -0.302 & \phantom{-}0.774 & \phantom{-}0.968 & \colorbox[HTML]{B2EAB1}{\textbf{\phantom{-}0.998}} & \colorbox[HTML]{B2EAB1}{\textbf{\phantom{-}0.911}} & \phantom{-}0.866 & \colorbox[HTML]{B2EAB1}{\textbf{\phantom{-}0.702}}\\
XCOMET-QE-Ensemble & -0.208 & \phantom{-}0.925 & \phantom{-}0.930 & \phantom{-}0.975 & \phantom{-}0.546 & \phantom{-}0.912 & \phantom{-}0.680\\
XLsimQE & \phantom{-}0.245 & -0.113 & \phantom{-}0.208 & \phantom{-}0.350 & -0.256 & -0.170 & \phantom{-}0.044 \\
   \bottomrule
    \end{tabular}}
    \caption{Results on the challenge sets where the good translation can only be identified through the source sentence. Upper block: reference-based metrics, lower block: reference-free metrics. Best results for each phenomenon and each group of models is marked in bold and green and the average over all can be seen in the last column.}
    \label{tab:src_disambig}
\end{table*}
\textbf{Finding: Reference-based metrics tend to ignore the source.}

In the \textsc{ACES} \textit{Mistranslation - Ambiguous Translations} category, examples were designed in such a way that given an ambiguous reference the correct translation candidate can only be identified through the source sentence (See an example in Table~\ref{tab:ACES_top_level_examples}). We leverage this property to present an analysis aimed at discovering how important the source is for different metrics. We exclude from the analysis all metrics that a) do not take the source and b) do not cover all language pairs. This leaves us with 22 metrics: seven reference-based metrics, fourteen reference-free metrics, and the \textsc{Random-sysname} baseline. In Table~\ref{tab:src_disambig} we present results for the \textit{Ambiguity - Discourse Connectives} (for the ambiguous English discourse connective ``since'' which can have either causal or temporal meaning), and \textit{Ambiguity - Occupation Names Gender} (male and female) challenge sets.

\begin{table}[h!]
    \centering
    \begin{tabular}{rc}
\toprule
    & \textbf{corr.}\\
    & \textbf{gain}\\
\midrule
Random-sysname & -0.052 \\ \midrule
COMET-22 & \phantom{-}0.042 \\
MetricX-23 & \phantom{-}0.004 \\
MetricX-23-b & -0.002 \\
MetricX-23-c & \phantom{-}0.008 \\
XCOMET-Ensemble & \textbf{\phantom{-}0.162} \\
XCOMET-XL & \phantom{-}0.110 \\
XCOMET-XXL & \phantom{-}0.016 \\ \midrule
cometoid22-wmt21 & \phantom{-}0.120 \\
cometoid22-wmt22 & \phantom{-}0.124 \\
cometoid22-wmt23 & \phantom{-}0.138 \\
CometKiwi & \phantom{-}0.454 \\
CometKiwi-XL & \phantom{-}0.148 \\
CometKiwi-XXL & \phantom{-}0.108 \\
GEMBA-MQM & \textbf{\phantom{-}1.107} \\
KG-BERTScore & \phantom{-}0.436 \\
MS-COMET-QE-22 & \phantom{-}0.198 \\
MetricX-23-QE & \phantom{-}0.272 \\
MetricX-23-QE-b & \phantom{-}0.296 \\
MetricX-23-QE-c & \phantom{-}0.142 \\
XCOMET-QE-Ensemble & \phantom{-}0.112 \\
XLsimQE & \phantom{-}0.184 \\
   \bottomrule
    \end{tabular}
    \caption{Results on the \textit{real-world knowledge commonsense challenge set} with reference-based metrics in the upper block and reference-free metrics in the lower block. The numbers are computed as the difference between the correlation with the subordinate clause in the source and the correlation without the subordinate clause in the source. Largest gains are bolded.}
    \label{tab:corr_gain}
\end{table}

In addition, we measure the correlation gain when metrics receive access to disambiguation information via the source -- for this we use the \textit{Real-world Knowledge - Commonsense} challenge set i.e. a scenario in which the source contains disambiguation information (See an example in Table~\ref{tab:ACES_examples}). In Table~\ref{tab:corr_gain} we observe that the correlation gain is lower for the majority of the reference-based metric correlation scores compared with the reference-free metric correlation scores, when access to the subordinate clause is provided via the source.

In line with last year's findings, we report that reference-based metrics still lag behind reference-free metrics in terms of their correlation on challenge sets that can only be disambiguated by looking at the source. This indicates that reference-based metrics still rely too much on the reference translation. We conclude that our initial finding from 2022 still holds: that reference-based metrics tend to ignore relevant information in the source. One exception is \textsc{XCOMET-Ensemble}, a reference-based metric that reaches similar correlations and correlation gains as some of the mid-performing reference-free metrics. We suspect that by training the same model to exhibit reference-based and reference-free behaviour, the model learns to utilise the information from the source in addition to the reference, when provided.


\subsubsection{How much do metrics rely on
surface overlap with the reference?}
\label{analysis:reference}
\textbf{Finding: Reference-based metrics still rely on reference overlap.}

Surface-level metrics are often too reliant on overlap with the reference. We aim to discover whether neural reference-based metrics submitted to the 2023 shared task are able to avoid this problem.  Using the \textit{Hallucination - Numbers and Named Entities} challenge set we compared how well reference-based and reference-free metrics\footnote{Excluding surface-level overlap metrics (\textsc{BLEU}, \textsc{chrF}, \textsc{fp200spBLEU}, \textsc{Partokengram\_F}, \textsc{Tokengram\_F}).} on average can identify \textit{number} and \textit{named entity} mismatches. In these challenge sets, we perform both word-level and character-level edits (i.e. substitutions) to simulate the hallucination behaviour. In order to thoroughly understand the behaviour of metrics under such hallucination errors, we introduced three levels. The first, easiest level follows \citet{freitag-etal-2021-results} and applies a change to an alternative translation to form an incorrect translation. The second level uses an alternative translation that is lexically very similar to the reference as the good translation and applies a change to the reference to form an incorrect translation. The third, and hardest level, uses an alternative translation that is lexically very different from the reference as the good translation and applies a change to the reference to form an incorrect translation. In this way, our challenge set tests whether number and named entity differences can still be detected as the surface similarity between the two translation candidates decreases and the surface similarity between the incorrect translation and the reference increases.
See an example of the different levels below as taken from the dataset paper - 
\begin{small}
\vspace{0.5cm}
\begin{tabularx}{0.95\columnwidth}{lX}
     SRC (es): & Sin embargo, Michael Jackson, Prince y \textbf{Madonna} fueron influencias para el álbum. \\
     REF (en): & Michael Jackson, Prince and \textbf{Madonna} were, however, influences on the album. \\\\\hline\\
    Level-1 \cmark: & However, Michael Jackson, Prince, and \textbf{Madonna} were influences on the album. \\
    Level-1 \xmark: & However, Michael Jackson, Prince, and \textbf{Garza} were influences on the album. \\\\\hline\\
    Level-2 \cmark: & However, Michael Jackson, Prince, and \textbf{Madonna} were influences on the album. \\
    Level-2 \xmark: &  Michael Jackson, Prince and \textbf{Garza} were, however, influences on the album.\\\\\hline\\
    Level-3 \cmark: & The record was influenced by \textbf{Madonna}, Prince, and Michael Jackson  though. \\
    Level-3 \xmark: & Michael Jackson, Prince and \textbf{Garza} were, however, influences on the album.\vspace{0.35cm}
\end{tabularx}
\end{small}

\begin{figure}
    \centering
    \includegraphics[width=0.45\textwidth]{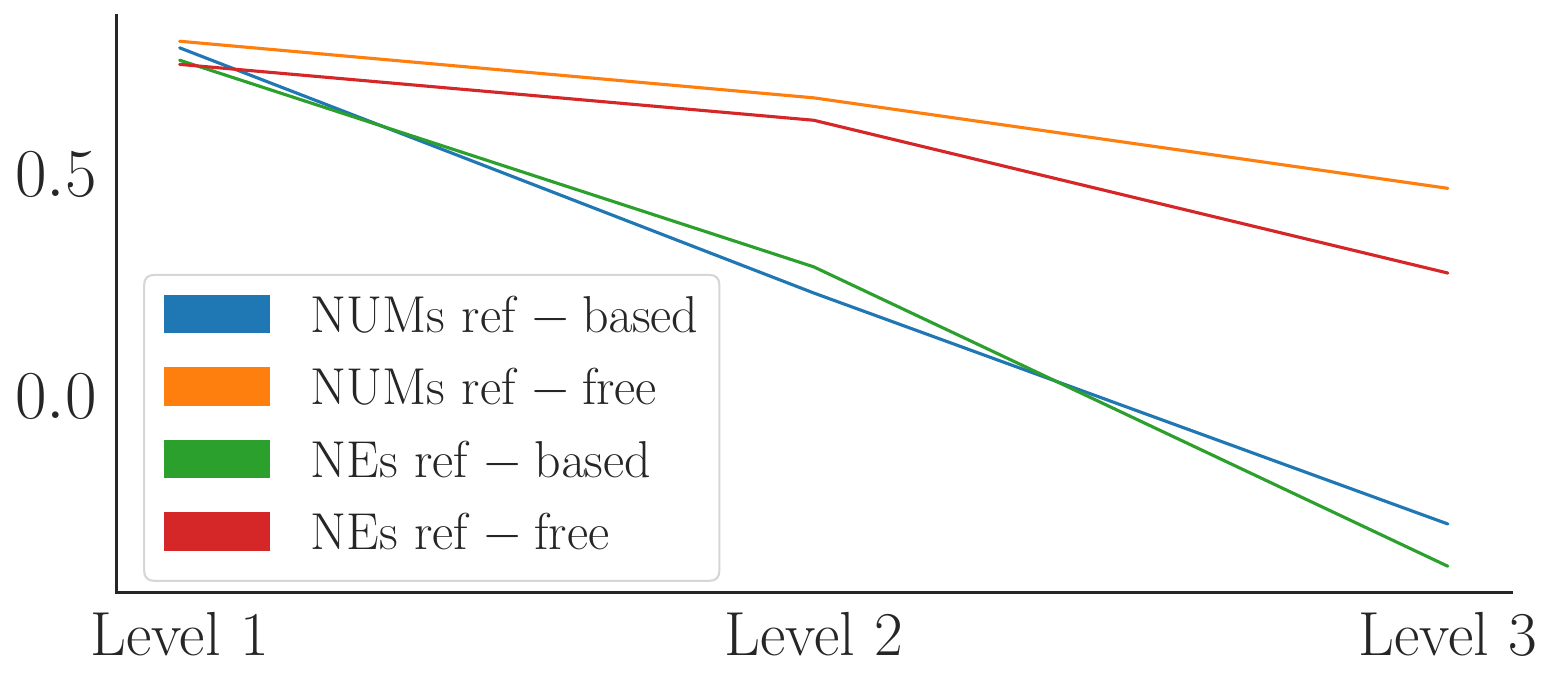}
    \caption{Decrease in correlation for reference-based and reference-free metrics on the named entity and number hallucination challenge sets.}
    \label{fig:corr_decrease}
\end{figure}

We take the average correlation for all reference-based and reference-free metrics that cover all languages. We then plot the decrease in correlation with increasing surface-level similarity of the incorrect translation to the reference (Figure~\ref{fig:corr_decrease}).
As in the corresponding analysis of the WMT 2022 metrics, we observe that, on average, reference-based metrics have a much steeper decrease in correlation than the reference-free metrics as the two translation candidates become more and more lexically diverse and the surface overlap between the incorrect translation and the reference increases. This indicates that reference-based metrics may prefer a) an incorrect translation in cases where it is lexically similar to the reference but contains a severe error over b) a good translation that shares little overlap with the reference.

We also observe a clear effect of surface-level overlap between the reference and the hypothesis on three challenge sets for which we have different versions of the good translation, where the error was corrected with: a) the corresponding \textit{correct token} from the reference and b) \textit{a synonym for the correct token} from the reference. In Table~\ref{tab:reference-overlap}, we can see a much larger difference in correlation between the challenge sets with reference-copied good translations and the challenge sets with the synonymous good translations, for the reference-based metrics as compared to the reference-free metrics. That is, it is much easier for reference-based metrics to identify mistranslations when the good translation matches a term in the reference compared with when a synonym is used. Furthermore, when the incorrect translation \textit{shares a high degree of lexical overlap with the reference but does not have the same meaning} (as in the \textit{Mistranslation - Lexical Overlap} challenge set based on adversarial paraphrase from PAWS-X \citep{yang-etal-2019-paws}), the reference-based metrics only reach a correlation of 0.05 ± 0.16 on average. In contrast, the reference-free metrics reach a correlation of 0.27 ± 0.16.

We again conclude that although state-of-the-art reference-based MT evaluation metrics are no longer solely reliant on surface-level overlap, it still has a considerable influence on their predictions.

\begin{table}[t!]
    \centering
    \small
    \begin{tabular}{ccc}
    \toprule
         & reference-based & reference-free \\
        \midrule
        \hyperref[subsec:real_hallucination]{hallucination} & -0.32 ± 0.15 & +0.06 ± 0.06 \\
        \hyperref[subsec:real_overly_literal]{overly-literal} & -0.22 ± 0.14
        & 0.00 ± 0.03\\
        \hyperref[subsec:real_untranslated]{untranslated} & -0.44 ± 0.11 & -0.03 ± 0.06\\
    \bottomrule
    \end{tabular}
    \caption{Average correlation difference and standard deviation between the challenge sets with reference-copied good translations and the challenge sets with the synonymous good translations.}
    \label{tab:reference-overlap}
\end{table}


\subsubsection{Do multilingual embeddings help design better metrics?}
\label{analysis:multilingual}

\textbf{Finding: Multilingual embeddings can be harmful with poor design.}

We are interested in the extent to which the representations in neural MT evaluation metrics, which are trained on multilingual models, are language-dependent. For this analysis, we investigated the effect of alignment of multilingual embeddings (including LLMs) on the evaluation task through the \textit{wrong-language} and \textit{untranslated - full sentences} phenomena for those metrics that provided scores for examples in all language pairs.
In the \textit{wrong-language} phenomenon, the incorrect translation contains a high-quality translation of the source in a similar language to the target language while the good translation is the machine translation output of the source sentence in the target language. In the challenge set for \textit{untranslated - full sentences}, the incorrect translation is a copy of the source sentence and the good translation is machine translation output in the target language.
Multilingual embeddings learn cross-lingual representations by reducing the language-specific properties during pretraining \citep{wu-dredze-2019-beto}. We hypothesised that making representations language agnostic may harm MT evaluation in cases where translations are extremely poor, such that they remain untranslated or hallucinate from a similar language.

\begin{figure}
    \centering
    \includegraphics[width=0.48\textwidth]{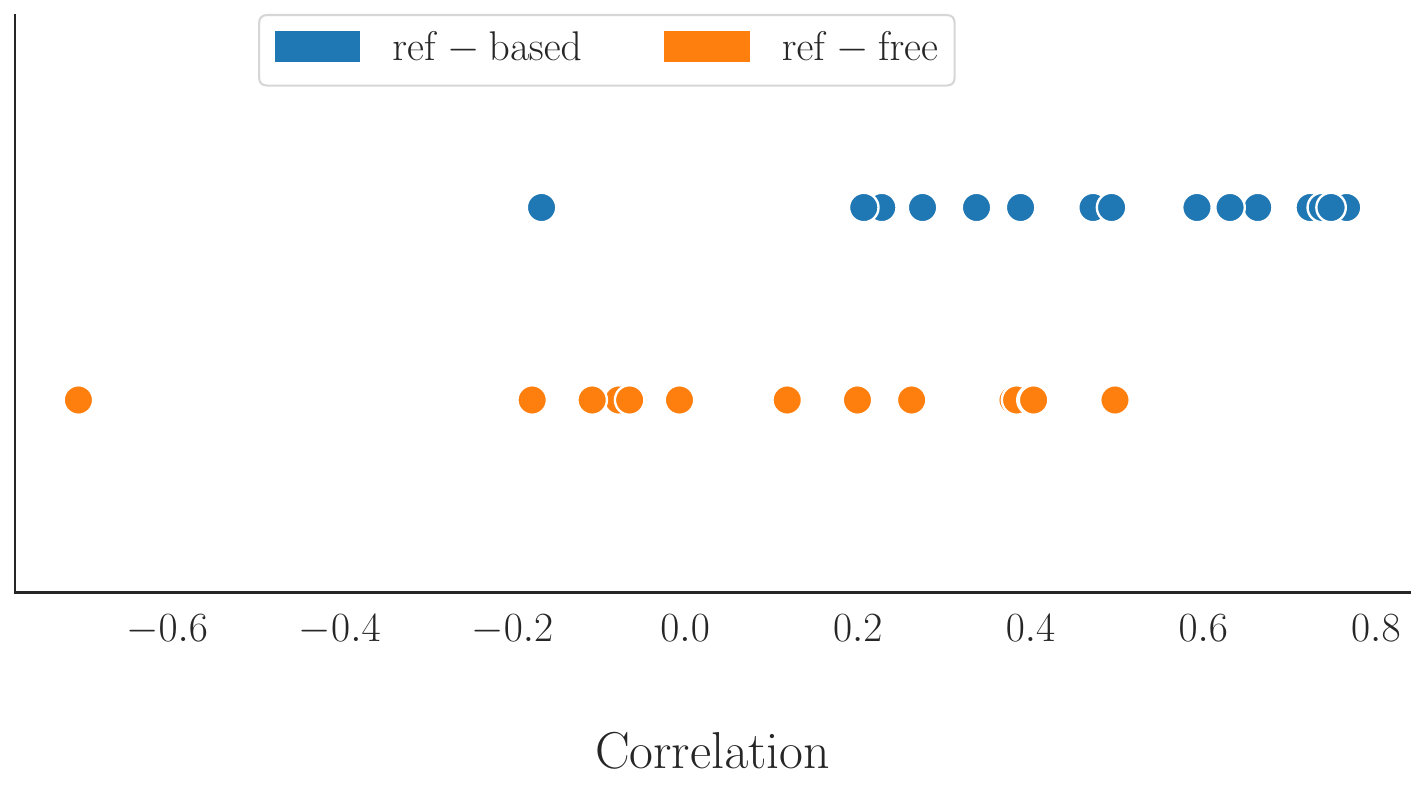}
    \caption{Correlation of reference-based metrics (blue) and reference-free metrics (orange) on the sentence-level untranslated test challenge set.}
    \label{fig:corr_copy_src}
\end{figure}

In Figure~\ref{fig:corr_copy_src} we plot the correlations for all reference-based and reference-free metrics. Overall, we observe that several metrics from 2023 have much better correlation scores than 2022 indicating that newer models have developed strategies to avoid learning language-agnostic representations. In particular, we find that many of the reference-free metrics submitted to the 2023 shared task have improved on the \textit{untranslated - full sentences} category (though a few reference-free metrics from 2022 had performance closer to 1, which is not the case with the 2023 metrics). This is a welcome change as we expect these metrics to perform a more faithful evaluation when many of the words remain untranslated in the hypothesis, especially in the lower resource setting. Whilst some reference-free metrics struggle considerably on this challenge set and almost always prefer the copied source to the real translation, reference-based metrics generally exhibit good correlation i.e. they can identify the copied source quite easily. As reference-based metrics tend to ignore the source, the scores are likely based on the similarity between the reference and the MT output. This is evident from their poor performance on the \textit{wrong - language} category (see Table~\ref{tab:analysis_overview}). This suggests that language-agnostic representations present in the multilingual space may harm performance.

\subsection{Training data size effects}
\label{subsec:data_effects}
\begin{figure}
    \centering
    \includegraphics[width=0.48\textwidth]{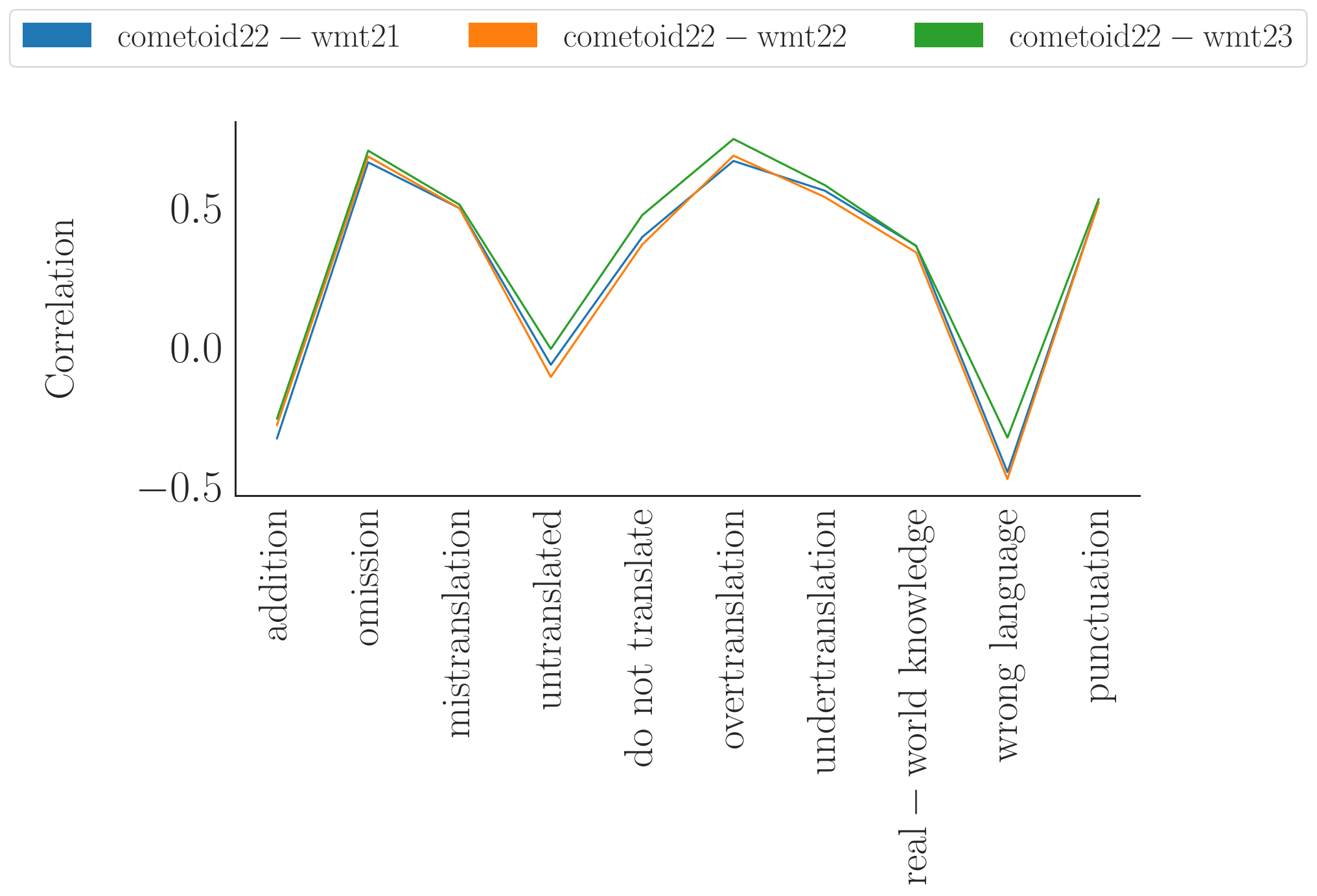}
    \caption{Correlations for different top-level phenomena categories with different models trained on successively more data.}
    \label{fig:data_effects}
\end{figure}

One submission this year, namely \textsc{Cometoid22}, submitted three different reference-free metric versions, each trained on successively more data. This allows us to investigate the effects of the metric training data size\footnote{Note that for \textsc{Cometoid22} this is not human judgement labelled data but rather pseudo labelled data where labels come from the reference-based COMET-22 model.} on the performance on ACES. (Note that we cannot draw any conclusions about the training data size of the pretraining models that are used.) In Figure~\ref{fig:data_effects} we can see the effect of training data size on the performance on the top-level phenomena categories. \textsc{Cometoid22-wmt23}, the model that has seen the most data, outperforms the other two metrics on almost all top-level categories. The correlation gain is especially pronounced for the \textit{untranslated}, \textit{do not translate} (content in the source is erroneously translated into the target language), \textit{overtranslation} (the target translation contains more specific information than the source) and \textit{wrong language} categories (see Table~\ref{tab:ACES_top_level_examples} for examples for each of the phenomena). For clearer insights as to where the performance gain comes from, we would need to analyse the training data in depth. However, it is evident from these results that more training data is beneficial for metric development. In the next section, we look at metric score changes over metric implementation cycles - where likely more than just the training data changed.

\subsection{Changes over one year}
\label{subsec:results_progress}
We compare the results of metrics submitted by the same teams last year and this year in Table~\ref{tab:analysis_2022_2023_delta}. 

\begin{table*}[h!]
\centering
\small
\begin{tabular}{@{}lcccccc@{}}
\toprule
 & \multicolumn{2}{c}{COMETKiwi} & KG-BERTScore & \multicolumn{3}{c}{XCOMET} \\
\cmidrule{2-3}\cmidrule{5-7}
                     & -XL & -XXL & & -Ensemble & -XL & -XXL \\
\midrule
addition             & -0.120            & -0.004             & -0.251            & \phantom{-}0.595           & \phantom{-}0.455     & \phantom{-}0.142      \\
omission             & -0.004            & -0.002             & \phantom{-}0.103             & \phantom{-}0.118           & -0.126    & -0.254     \\
mistranslation       & -0.005            & \phantom{-}0.013              & \phantom{-}0.077             & \phantom{-}0.126           & \phantom{-}0.038     & \phantom{-}0.005      \\
untranslated         & \phantom{-}0.000             & \phantom{-}0.142              & \phantom{-}0.266             & -0.181          & -0.342    & -0.362     \\
do not translate     & -0.395            & -0.553             & \phantom{-}0.000             & \phantom{-}0.053           & \phantom{-}0.079     & -0.105     \\
overtranslation      & \phantom{-}0.027             & \phantom{-}0.035              & \phantom{-}0.119             & \phantom{-}0.073           & -0.067    & \phantom{-}0.017      \\
undertranslation     & -0.019            & -0.021             & \phantom{-}0.077             & \phantom{-}0.014           & -0.132    & -0.025     \\
real-world knowledge & -0.020            & \phantom{-}0.100              & \phantom{-}0.107             & \phantom{-}0.003           & -0.123    & -0.198     \\
wrong language       & -0.014            & -0.173             & -0.618            & -0.296          & -0.232    & -0.395     \\
punctuation          & -0.037            & \phantom{-}0.004              & \phantom{-}0.264             & \phantom{-}0.206           & -0.144    & \phantom{-}0.006      \\
\midrule
ACES-Score           & -1.04\phantom{1}             & -0.38\phantom{1}               & \phantom{-}0.40\phantom{1}               & 4.23\phantom{1}             & \phantom{-}0.21\phantom{1}       & -1.64\phantom{1}      \\
\bottomrule
\end{tabular}
\caption{Comparison of average Kendall’s tau-like correlation: delta calculated as 2023 score minus 2022 score.}
\label{tab:analysis_2022_2023_delta}
\end{table*}

We report changes in performance in terms of deltas, computed by subtracting the 2022 score from the 2023 score. We do this for the following pairs of metrics: \textsc{KG-BERTScore} (2022) and \textsc{KG-BERTScore} (2023); \textsc{COMETKiwi} (2022) paired with \textsc{COMETKiwi-XL} (2023) and \textsc{COMETKiwi-XXL} (2023); \textsc{COMET-22} (2022) paired with \textsc{XCOMET-Ensemble} (2023), \textsc{XCOMET-XL} (2023) and \textsc{XCOMET-XXL} (2023). 

We observe that \textsc{KG-BERTScore} has improved over its performance of last year. From the description provided by the metric developers, the main difference is that the 2023 version of \textsc{KG-BERTScore} metric uses COMET-QE instead of BERTScore \citep{DBLP:conf/iclr/ZhangKWWA20} to compute the similarity between the source and the hypothesis. Whilst we might therefore attribute the increase in performance to this change, a more systematic comparison of the two metric versions would be required to confirm whether this is the only contributing factor. 

The metrics in the \textsc{COMETKiwi} family exhibit: a slight drop in performance (\textsc{COMETKiwi-XL}) and a similar performance to that of last year (\textsc{COMETKiwi-XXL}). The difference can be attributed to changing the underlying encoder, XLM-R XL and XLM-R XXL \citep{DBLP:journals/corr/abs-2105-00572} respectively, and the use of additional fine-tuning data made available this year. We have seen that the addition of more training data helps in Section~\ref{subsec:data_effects}. Considering that there is no improvement in the performance, we question if an increase in the underlying model capacity of the encoder alone is useful for obtaining better MT evaluation.

Performance change for the XCOMET family is variable: there is a performance increase for \textsc{XCOMET-Ensemble} (compared to COMET-22), for \textsc{XCOMET-XL} the increase is smaller, and the performance of \textsc{XCOMET-XXL} is degraded. The XCOMET family is designed to provide both a quality score and an error span. Considering that the metric also provides an explanation of the scores without hurting the performance, this is indeed a positive change.
Finally, it is worth noting that for \textit{all} metrics in Table~\ref{tab:analysis_2022_2023_delta} a change in performance is observed for almost all \textsc{ACES} categories, for all metrics.

Whilst it is not possible to draw conclusions or make predictions about the future of metric development based solely on the observations from two consecutive metrics shared tasks, we highlight several high-level changes. Firstly, we note the participation of many more COMET-based metrics in 2023, compared with 2022. This is presumably based on the success of COMET at previous shared tasks and its adoption within the MT community. We find that three metrics from 2022 are now used as baseline metrics namely \textsc{COMET-22}, \textsc{CometKiwi}, and \textsc{MS-COMET-QE-22}. In contrast to the submissions in 2022, we find some new metrics that use lexical overlap through text matching or embeddings (\textsc{Tokengram\_F}, \textsc{Partokengram\_F}, and \textsc{eBLEU}). However, their performance trend is similar to other surface overlap metrics. This year has also seen submissions based on large language models (\textsc{embed\_llama} and \textsc{GEMBA-MQM}). As seen in Section~\ref{sec:aces_overview}, their moderate performance indicates the need for more effective approaches. Additionally, we note an overall increase between 2022 and 2023 in the number of metrics submitted to WMT that a) provide segment-level scores and b) provide scores for all language pairs and directions in ACES. There were 37 segment-level metrics at WMT 2022, 24 of which covered the language pairs and directions in ACES, compared with 47 and 33, respectively in 2023. This suggests that the interest in metric development remains high, and could be increasing.

From our analyses in Section~\ref{subsec:results_analysis}, we also draw similar conclusions to \citet{amrhein-etal-2022-aces} with the exception of reference-free metrics improving at the \textit{Untranslated - Full Sentences} task. Despite the success of LLMs across various tasks \citep{DBLP:journals/corr/abs-2005-14165}, leveraging them to evaluate translated outputs still requires some improved design strategies. All these observations suggest that evaluating MT outputs is indeed a hard problem \citep{neubig2022is}. While we do have a good suite of metrics to provide a proxy for evaluation, there are indeed several interesting challenges that need to be tackled before we find an ideal evaluation regime. And even then, we need to continuously monitor this to ensure that we do not optimise towards metric weaknesses that we have not yet discovered.



\subsection{Recommendations}
\label{subsec:recommendations}
We provide the same recommendations as last year:
\paragraph{No metric to rule them all.} There is no consistent winning metric across all categories (see Table~\ref{tab:analysis_overview}). We recommend developing evaluation methods that combine different design strategies for robust evaluation. We also recommend innovation in the ensemble building as simple strategies like majority voting do not lead to significant improvement \citep{moghe-etal-2023-extrinsic}.  We find that some submissions in this year's shared task already contain ensembles (\textsc{XCOMET-Ensemble, XCOMET-QE-Ensemble}) which suggests that our recommendations are in line with the efforts of the community.

\paragraph{The source matters.} The trend where reference-based metrics tend to disregard information in the source is also persistent, as seen in Section~\ref{subsec:results_analysis}. We also observe that reference-free metrics are highly competitive with reference-based metrics as seen in Table~\ref{tab:analysis_overview} and also in \citet{freitag-etal-2022-results, zerva-etal-2022-findings}, \textit{inter alia}.
Furthermore, as references are often not perfect themselves \citep{freitag-etal-2020-bleu}, it is ideal to develop evaluation regimens that focus more on the information in the source sentence than the references. 

\paragraph{Surface overlap still prevails.} Neural metrics were introduced to overcome surface-level overlap present in the string-based metrics. However, the results in Section~\ref{analysis:reference} suggest that neural metrics tend to focus more on lexical overlap than semantic content. We thus recommend including paraphrases in the training regime as well as designing loss functions that explicitly discourage surface-level overlap. 

\noindent Lastly, simple strategies to model language-specific information in the metrics could also improve the robustness of the metrics to language pair attacks.

\section{Conclusion}
\label{sec:conclusion}
We re-submitted the \textsc{ACES} Challenge Set to WMT2023 to identify the strengths and weaknesses of the metrics submitted to this year's shared task. Overall, we find similar trends to that of last year. While neural metrics tend to be better, different categories of metrics have different strengths, and we do not find one clear winner. With respect to the metrics that were resubmitted with some design changes, we find that these design changes have variable outcomes with a performance drop in some cases.
The major challenges of (i) metrics not paying enough attention to the source, (ii) reference-based metrics still relying on surface-level overlap, and (iii) over-reliance on multilingual embeddings still persist. Hence, our recommendations are also similar to that of last year: build ensembles of different design families, encourage development that better utilises information in the source, include diverse training examples to reduce the influence of surface-level overlap, and carefully determine the influence of multilingual embeddings/LLMs on MT evaluation. 

\section*{Limitations}
When comparing the results of the baseline metrics common to the 2022 and 2023 metrics shared tasks, we observed differences in the scores returned for a small subset (2,659; approx 7\%)  of the ACES examples. A subsequent investigation suggested that differences in the pre-processing steps by the shared task organisers in 2022 and 2023 may have led to the differences; we further conjecture that differences in handling the double quotes present in some of the ACES examples may be one of the main causes. Regardless of the source of the differences, we highlight that care should be taken when pre-processing the ACES dataset prior to benchmarking metric performance, especially when the aim is to draw comparisons with results reported in previous work. However, we note that this issue is not specific to ACES, but may potentially affect any text-based dataset. With the exception of the comparison of results from 2022 and 2023 in Section~\ref{subsec:results_progress}, for which we used the subset of 33,817 examples which were unaffected by pre-processing differences, all other results reported in this paper use the full set of 36,476 ACES examples. We also note that ideally, incorrect processing of double quotes by a metric should not lead to a difference in scores especially when dealing with semantic errors.

As we re-submitted the same version of the \textsc{ACES} dataset to WMT 2023, the same biases described in \citet{amrhein-etal-2022-aces} remain: 1) there is greater coverage in terms of phenomena and number of examples for some language pairs (particularly en-de and en-fr), 2) more examples are provided for categories for which examples may be generated automatically, compared to those that required manual construction/filtering, 3) errors present in the datasets used to construct the examples may have propagated through into \textsc{ACES}, 4) the focus of the \textsc{ACES} is on accuracy errors; the inclusion and evaluation of fluency errors remains a direction for future work.

\textsc{ACES} consists of examples that target a range of linguistic phenomena, which are then arranged in a hierarchy of error categories. In order to provide metric profiles over this range of error categories we require segment-level scores. We therefore report only results for those metrics submitted to WMT 2023 that provide segment-level scores; metrics that provide only system-level outputs are excluded. Further, we excluded those metrics that did not provide scores for all of the language pairs in \textsc{ACES} from the results and analyses in this paper.

The 2023 WMT metric shared task evaluated metrics at the paragraph level for English-German. Currently, ACES is not able to capture document-level metric performance. We hope such challenge sets will become available in the near future to be able to track metric improvements beyond the sentence level.

\section*{Ethics Statement}
As described in \citet{amrhein-etal-2022-aces} some examples within the ACES challenge set exhibit biases. However, this is necessary in order to expose the limitations of existing metrics. The challenge set is already publicly available.

\section*{Acknowledgements}
We thank the organisers of the WMT 2023 Metrics task for organising the Challenge Sets shared task, and the shared task participants for scoring our challenge sets with their systems. We thank Nikolay Bogoychev and the anonymous reviewers for their insightful comments and suggestions. This work was supported in part by the UKRI Centre for Doctoral Training in Natural Language Processing, funded by the UKRI (grant EP/S022481/1) and the University of Edinburgh (Moghe), and by the Swiss National Science Foundation (project MUTAMUR; no. 176727) (Amrhein). We also thank Huawei (Moghe) and the RISE Center for Applied AI (Guillou) for their support.

\bibliography{anthology,custom}
\bibliographystyle{acl_natbib}

\appendix
\section{Examples from ACES}
\label{sec:top_category}
We shall now list one example from every top-level category in Table~\ref{tab:ACES_top_level_examples}. We reuse most of the examples mentioned in the original paper under the respective categories. 

\begin{table*}[ht!]
    \small
    \centering
    \resizebox{\textwidth}{!}{
    \begin{tabular}{rl}
        \toprule
        & \textbf{Addition}\\
        & \textit{target includes content not present in the source}\\\\
        SRC (de): & In den letzten 20 Jahren ist die Auswahl in Uptown Charlotte exponentiell gewachsen.\\
        REF (en): & In the past 20 years, the amount in Uptown Charlotte has grown exponentially.\\
        \cmark: & Over the past 20 years, the selection in Uptown Charlotte has grown exponentially.\\
        \xmark: & Over the past 20 years, the selection of \textbf{child-friendly options} in Uptown Charlotte has grown exponentially.\\
        \midrule
        
        & \textbf{Omission}\\
        & \textit{errors where content is missing from the translation that is present in the source}\\\\
        SRC (fr): & Une tornade est un tourbillon d'air à basse-pression en forme de colonne, l'air alentour est aspiré vers l'intérieur et le haut.\\
        REF (en): & A tornado is a \textbf{spinning column} of very low-pressure air, which sucks the surrounding air inward and upward.\\
        \cmark: & A tornado is a \textbf{column-shaped} low-pressure air turbine, the air around it is sucked inside and up.\\
        \xmark: & A tornado is a low-pressure air turbine, the air around it is sucked inside and up.\\
        \midrule
        
        & \textbf{Untranslated - Word Level} \\
        & \textit{errors occurring when a text segment that was intended for translation is left untranslated in the target content} \\\\
      SRC (fr): & À l'origine, l'émission mettait en scène des \textbf{comédiens de doublage} amateurs, originaires de l'est du Texas. \\
     REF (de): & Die Sendung hatte ursprünglich lokale Amateur\textbf{synchronsprecher} aus Ost-Texas. \\
     \cmark{} (copy): & Ursprünglich spielte die Show mit Amateur\textbf{synchronsprechern} aus dem Osten von Texas. \\
     \cmark{} (syn.): & Ursprünglich spielte die Show mit Amateur-\textbf{Synchron-Schauspielern} aus dem Osten von Texas. \\
     \xmark: & Ursprünglich spielte die Show mit Amateur-\textbf{Doubling-Schauspielern} aus dem Osten von Texas. \\
        \midrule
        
        & \textbf{Mistranslation - Ambiguous Translation} \\
        & \textit{an unambiguous source text is translated ambiguously} \\\\
        SRC (de): &  Der Manager feuerte \textbf{die} Bäcker\textbf{in}. \\
        REF (en): & The manager fired the baker. \\
        \cmark: & The manager fired the \textbf{female} baker. \\
        \xmark: & The manager fired the \textbf{male} baker. \\
        \midrule

        & \textbf{Do Not Translate} \\
        & \textit{content in the source that should be copied to the output in the source language, but was mistakenly translated into the target language.} \\\\

       SRC (en): & Dance was one of the inspirations for the exodus - song \textbf{``The Toxic Waltz''}, from their 1989 album ``Fabulous Disaster''. \\
     REF (de): & Dance war eine der Inspirationen für das Exodus-Lied \textbf{„The Toxic Waltz“} von ihrem 1989er Album „Fabulous Disaster“. \\
     \cmark: & Der Tanz war eine der Inspirationen für den Exodus-Song \textbf{„The Toxic Waltz“}, von ihrem 1989er Album „Fabulous Disaster''. \\
     \xmark: & Der Tanz war eine der Inspirationen für den Exodus-Song \textbf{„Der Toxische Walzer“}, von ihrem 1989er Album „Fabulous Disaster''.\\
        \midrule

        & \textbf{Undertranslation} \\
        & \textit{erroneous translation has a meaning that is more generic than the source} \\\\

        SRC (de): & Bob und Ted waren Brüder. Ted ist der \textbf{Sohn} von John. \\
        REF (en): & Bob and Ted were brothers. Ted is John's \textbf{son}. \\
        \cmark: & Bob and Ted were brothers, and Ted is John's \textbf{son}. \\
       \xmark: & Bob and Ted were brothers. Ted is John's \textbf{male offspring}. \\
        \midrule
        
        & \textbf{Overtranslation} \\
        & \textit{erroneous translation has a meaning that is more specific than the source} \\\\

        SRC (ja): & \begin{CJK}{UTF8}{min}その 40 分の\textbf{映画}はアノーがアラン・ゴダードと協力して脚本を書いた。\end{CJK} \\
        REF (en): & The 40-minute \textbf{film} was written by Annaud with Alain Godard. \\
        \cmark: &The 40-minute \textbf{film} was written by Annaud along with Alain Godard. \\
       \xmark: & he 40-minute \textbf{cinema verite} was written by Annaud with Alain Godard. \\
        \midrule

         & \textbf{Real-world Knowledge - Textual Entailment} \\
         & \textit{meaning of the source/reference is entailed  by the ``good'' translation} \\\\
         SRC (de): & Ein Mann \textbf{wurde ermordet}.\\
         REF (en): & A man \textbf{was murdered}.\\
         \cmark: & A man \textbf{died}.\\
         \xmark : & A man \textbf{was attacked}.\\
         \midrule    
        
         & \textbf{Wrong Language} \\
         & \textit{incorrect translation is a perfect translation in a related language} \\\\
          SRC (en): & Cell comes from the Latin word cella which means small room. \\
         REF (es): & El término célula deriva de la palabra latina cella, que quiere decir «cuarto pequeño». \\
         \cmark\ (es): & La célula viene de la palabra latina cella que significa habitación pequeña. \\
         \xmark\ (ca): & Cèl·lula ve de la paraula llatina cella, que vol dir habitació petita. \\

         \bottomrule
    \end{tabular}}
    \caption{Examples from each top-level accuracy error category in \textsc{ACES}. An example consists of a source sentence (SRC), reference (REF), good (\cmark) and incorrect (\xmark) translations, language pair, and a phenomenon label. We also provide a description of the relevant phenomenon. en: English, de: German, fr: French, ja: Japanese, es: Spanish, ca: Catalan}
    \label{tab:ACES_top_level_examples}
\end{table*}

\end{document}